\documentclass[11pt,a4paper]{article}
\usepackage[affil-sl,noblocks]{authblk}
\usepackage[hyperref]{acl2020}
\usepackage{times}
\usepackage{latexsym}
\usepackage{mdframed}
\usepackage{graphicx}
\usepackage{tikzsymbols}

\usepackage{microtype}
\usepackage{amsmath}
\usepackage[absolute,showboxes,overlay]{textpos}
\TPMargin{3pt}

\aclfinalcopy

\title{FHAC at GermEval 2021: Identifying German toxic, engaging, and fact-claiming comments with ensemble learning}

\author[1]{Tobias Bornheim}
\author[1]{Niklas Grieger}
\author[1,2,*]{Stephan Bialonski}
\affil[1]{Department of Medical Engineering and Technomathematics\authorcr
FH Aachen University of Applied Sciences, Jülich, Germany\authorcr}
\affil[2]{Institute for Data-Driven Technologies\authorcr
FH Aachen University of Applied Sciences, Jülich, Germany}
\affil[*]{\textit{bialonski@fh-aachen.de}}

\date{}

\begin{document}
\maketitle
\begin{abstract}
The availability of language representations learned by large pretrained neural network models (such as BERT and ELECTRA) has led to improvements in many downstream Natural Language Processing tasks in recent years. Pretrained models usually differ in pretraining objectives, architectures, and datasets they are trained on which can affect downstream performance.
In this contribution, we fine-tuned German BERT and German \mbox{ELECTRA} models to identify toxic (subtask 1), engaging (subtask 2), and \mbox{fact-claiming} comments (subtask 3) in Facebook data provided by the GermEval 2021 competition. We created ensembles of these models and investigated whether and how classification performance depends on the number of ensemble members and their composition.
On \mbox{out-of-sample} data, our best ensemble achieved a macro-F1 score of $0.73$ (for all subtasks), and F1 scores of $0.72$, $0.70$, and $0.76$ for subtasks 1, 2, and 3, respectively.
\end{abstract}

\begin{textblock*}{17cm}[-0.1,0](0cm,27.4cm)
    \centering
    \small
    This work was published as part of the conference proceedings of the GermEval 2021 Workshop on the Identification of Toxic, \\Engaging, and Fact-Claiming Comments available online at DOI: \href{https://dx.doi.org/10.48415/2021/fhw5-x128}{10.48415/2021/fhw5-x128}. \\
    Please cite as: Tobias Bornheim, Niklas Grieger, and Stephan Bialonski. FHAC at GermEval 2021: Identifying German toxic, \\engaging, and fact-claiming comments with ensemble learning. In \textit{Proc. GermEval 2021 Workshop on Identification of Toxic, \\Engaging, and Fact-Claiming Comments: 17th KONVENS 2021}, pages 105--111, Online (2021).
\end{textblock*}

\section{Introduction}
Social media plays a role in the spreading of problematic content, ranging from conspiracy theories and concerted misinformation campaigns to offensive language in user comments~\citep{Zhuravskaya2020}. Moderating comments remains a challenge due to the ever-increasing amount of user-generated content created daily. One promising approach to addressing this challenge are techniques from Natural Language Processing (NLP) that support manual moderation processes by, for example, alerting human moderators to potentially problematic comments.

% 2 Progress in NLP: (i) methodological and (ii) availability of annotated data (-> shared tasks)
Among the many factors that have driven recent progress in NLP, we note in particular (i) methodological advances in language modeling and (ii) the availability of annotated data due to shared tasks. Recent methodological advances can be traced back to the invention and availability of deep neural network models. A major contribution was the invention of the transformer architecture, which harnesses self-attention mechanisms to effectively model long-range correlations in series of tokens (e.g., sentences) \citep{Vaswani2017}. Based on the transformer architecture, neural network models such as BERT \citep{Devlin2018,Rogers2020} were proposed and trained in a self-supervised fashion on large unannotated text corpora. Language representations learned by BERT turned out to be effective in many downstream tasks, leading to new state-of-the-art NLP systems. While \emph{masked language modeling} and \emph{next sentence prediction} are used as objectives in self-supervised pretraining to learn representations in BERT, other pretraining objectives such as \mbox{\emph{replaced token detection}} (ELECTRA, \citet{Clark2020}) have been demonstrated to yield language representations that can be better suited for various downstream tasks~\citep{Xia2020}. Furthermore, language representations have been learned in multilingual language models (such as mBERT) and in language specific BERT models \citep{Nozza2020}. Recent German specific language models include the BERT-based models GBERT \citep{Chan2020} and GottBERT \citep{Scheible2021} as well as the ELECTRA based  model GELECTRA \citep{Chan2020}.

The second factor driving progress in NLP has been recurring shared tasks that foster the exchange of ideas, the development and comparative assessment of methods, as well as the availability of annotated data~\citep{Nissim2017}. In addition to multilingual shared task campaigns (see, e.g., \citet{Mandl2019,Basile2019}), there exist language-specific shared task evaluations such as GermEval which focus on NLP for the German language. A series of GermEval tasks addressed the challenge of reliably identifying offensive language \citep{Wiegand2018} and distinguishing between profane, offensive, or abusive language found in Twitter tweets \citep{Struss2019}. The GermEval 2021 shared task on identifying  toxic, engaging, and fact-claiming comments \citep{Risch2021} provided German comments from a Facebook page of a political talk show of a German television broadcaster.

In this contribution, we investigate the ability of ensembles of GBERT and GELECTRA models to identify toxic, engaging, and fact-claiming comments. Our work was inspired by previous studies on German BERT models \citep{Graf2019} and ensemble approaches \citep{Risch2018,Risch2020}. We study the dependence of classification performance on the number of ensemble members and ensemble composition. Finally, we describe the models that were evaluated in the \mbox{GermEval 2021} shared tasks and report performance scores achieved on out-of-sample data.
The implementation details of our experiments are available online\footnote{\url{https://github.com/fhac-fb9-ds/germeval2021}}.

\section{Data and tasks}

\begin{figure}
    \begin{mdframed}{
    Frau Barley war mit ihrem dummdreisten überheblichen Grinsen wirklich nicht zu ertragen. (TOXIC)\\
    \\
    Da dreht sich jemand im Kreis. Die 7 Prozent kann der Vermieter doch auf die Miete schlagen. Diese starke Position ergibt sich durch den Markt (Angebot und Nachfrage), da ist es egal. [...] (ENGAGING, FACT-CLAIMING)} 
    \end{mdframed}
    \caption{Samples (Facebook comments) from the dataset of the GermEval 2021 shared task. }
    \label{fig:example_data}
\end{figure}

The dataset of the shared task consisted of 3244 annotated Facebook comments and was provided by the organizers of GermEval 2021~\citep{Risch2021}. The comments were drawn from a Facebook page of a political talk show of a German television broadcaster from February till July 2019 and were anonymized by replacing links to users by @USER, links to the show by @MEDIUM, and links to the moderator of the show by @MODERATOR. Four trained annotators labeled the data by three categories, indicating toxic, engaging, and \mbox{fact-claiming} comments (see figure~\ref{fig:example_data}).

The shared task consisted of three binary classification subtasks that aimed at predicting whether a given comment belonged to a category (class) or not~\citep{Risch2021}. Comments were considered toxic (subtask 1) when they could violate the rules of polite behavior or violated democratic discourse values. Automated identification of such comments can be particularly valuable for managers of online communities. Comments were considered engaging (subtask 2) when they were in line with deliberative principles such as rationality, reciprocity, and mutual respect. Such comments might encourage user engagement and could be made more visible in online communities. Finally, comments were considered fact-claiming (subtask 3) if they contained assertion of facts and/or provided evidence by citing external sources. Identifying such comments may constitute a preprocessing step that could assist community managers to filter out misinformation and fake news.

\begin{figure}
    \includegraphics[width=\linewidth]{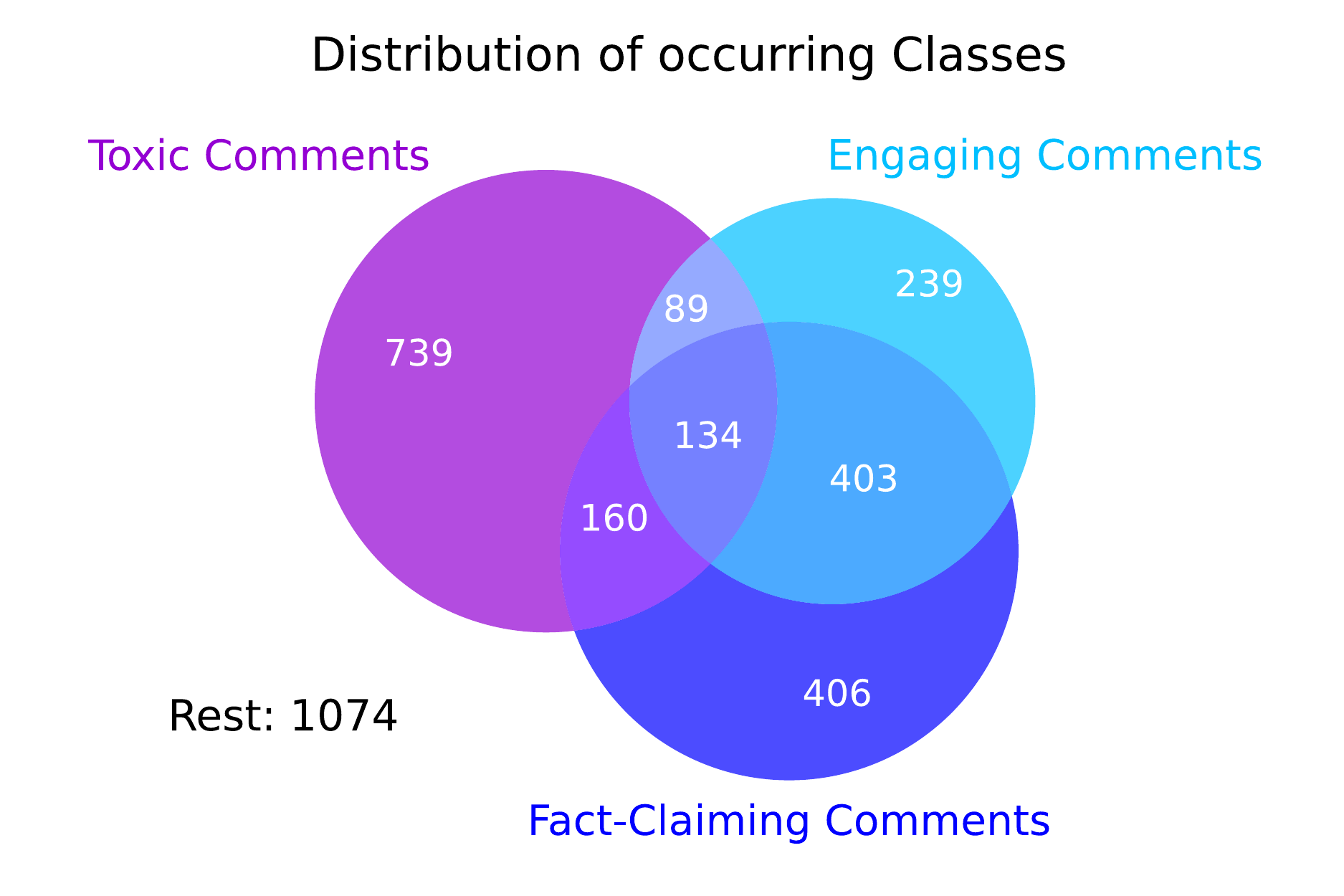}
    \caption{Venn diagram showing the numbers of comments that were labeled as toxic, engaging, or \mbox{fact-claiming}. 33\,\% of all comments were not assigned any class, whereas 24\,\% were attributed to more than one class.}
    \label{fig:venn-diagram}
\end{figure}

We did not observe any major class imbalance~\cite{Haixiang2017} as all three classes occurred with similar frequencies in the dataset (35\% toxic, 27\% engaging, 34\% fact-claiming).
However, the Venn diagram (see figure~\ref{fig:venn-diagram}) demonstrated significant overlap between classes where 24\% of all comments were attributed to more than one class. For instance, the large overlap between the engaging and the fact-claiming classes may point towards a correlation between these two classes. Such label correlations can be exploited by \mbox{\emph{multi-label classification}} approaches to improve classification performance~\citep{Zhang2014}. Thus we pursued a two-fold strategy. (i) In our first approach, we trained a multi-label classifier to predict all the possible class attributions for a given comment. Such models are called \emph{multi-label} in the following. (ii) In our second approach, we trained separate binary classifiers that aimed at distinguishing between toxic and non-toxic, engaging and non-engaging, or fact-claiming and non-fact-claiming classes, respectively. This approach of transforming a multi-label classification task into multiple single-label classification tasks is also known as a \emph{binary relevance transformation} \citep{Zhang2014}. We call such models \emph{single-label} in the following. Note that in this case, training a size 30 ensemble to classify comments means training three separate size 30 ensembles, each making predictions for one of the three binary classification tasks.

\section{Methods}

\subsection{Preprocessing and data splits}
\label{ssec:preprocessing}

\emph{Preprocessing.} All data (i.e., training and test data) was preprocessed as follows. 
First, all duplicates in the training data were removed, reducing the 3244 training samples to 3226 unique samples.
In a second step, all in-word whitespaces were removed (e.g. transforming the sequence ``A~K~T~U~E~L~L~!'' into the word ``AKTUELL!'') \cite{Paraschiv2019}.
Third, emojis were buffered with additional whitespaces such that words immediately followed by an emoji were not tokenized as unknown and emojis were tokenized separately (e.g., transforming the sentence ``I always start my day with a \mbox{coffee\hspace{2pt}\Coffeecup\Coffeecup\Coffeecup}'' into ``I always start my day with a coffee \hspace{3pt} \Coffeecup \hspace{3pt} \Coffeecup \hspace{3pt} \Coffeecup'') \cite{Risch2020}.
Fourth, any leading, trailing or consecutive whitespaces were removed.
Last, all comments were limited to a maximum length of 200 tokens to save computational resources and speed up training.
Only 49 out of the 3226 unique sentences in the training data and 21 out of 944 sentences in the test data were affected by this step.

\emph{Data splits.} During model exploration, models were trained with a 5-fold cross validation scheme (i.e., with 5 folds, each containing 20\% of the randomly shuffled training data). The final models evaluated by GermEval 2021 were trained on all training data (i.e., on all folds) to optimize model fitting.
Furthermore, during model exploration as well as for the final models, 10\% of the data in the training folds was randomly selected to act as an \emph{early stopping set} (see section~\ref{sec:training}) that was not used for training.

\subsection{Models}
\label{ssec:models}

We studied two recent transformer-based German language models \cite{Chan2020} called GBERT, based on the BERT architecture \cite{Devlin2018}, and GELECTRA, based on the ELECTRA architecture \cite{Clark2020}.
Both models use a tokenizer with a vocabulary size of 31k cased words.
From the different pretrained versions of these models, we chose gbert-large\footnote{\url{https://huggingface.co/deepset/gbert-large}} and gelectra-large\footnote{\url{https://huggingface.co/deepset/gelectra-large}}, both with a hidden states count of 1024.

A classification head was added on top of the first output vector of both pretrained transformer models. 
In the GBERT architecture, the mentioned output vector was generated by inserting a classification token at the beginning of every input sequence, which is used for the \mbox{\emph{next sentence prediction}} task during pretraining \cite{Devlin2018}.
The classification head consisted of a linear layer with the same hidden size as the transformer model, followed by a tanh activation function and another linear layer \cite{Wolf2020}.
Although the GELECTRA architecture does not use any \emph{next sentence prediction} task during pretraining \cite{Clark2020}, a classification token is still prepended to the transformer input and can be used during fine-tuning.
The classification head of GELECTRA had the same architecture as that of the GBERT model, except that a GELU activation \cite{Hendrycks2016} was used instead of a tanh activation \cite{Wolf2020}.

All linear layers of both classification heads were initialized randomly, except for the first layer of the GBERT classifier, which was initialized with the weights learned during the pretraining task.
Depending on whether the models were single-label or  multi-label classifiers, the final linear layer consisted of either two outputs followed by a softmax function or three outputs followed by a sigmoid function.

\subsection{Training}
\label{sec:training}

\emph{Evaluation scores.} To evaluate the prediction performance of a model, we determined the F1 score  following the definition used throughout the GermEval shared tasks \citep{Wiegand2021}. In GermEval, the F1 score of a binary classifier is determined by calculating precision and recall for the positive class (e.g., "toxic") and for the negative class (e.g. "non-toxic"). Precision and recall are then averaged over the two classes. The F1 score is calculated as harmonic mean over averaged recall and averaged precision. By taking the arithmetic mean of F1 scores of each binary classifier, we obtained the macro-F1 score $\overline{F1} = \frac{1}{3} (F1_{toxic} + F1_{engaging} + F1_{fact})$.
During model exploration, $\overline{F1}$ scores were determined for all five validation folds, and their mean and standard deviation were determined.
We considered a model to be superior to other models if its $\overline{F1}$ score averaged over all validation folds (of the cross validation) was larger than those of the other models.

\emph{Training scheme.} Each model (i.e., transformer with classification head) was trained with a batch size of 24 samples for 10 epochs using the AdamW optimizer \cite{Loshchilov2019}.
We used a learning rate of $\eta=5\cdot 10^{-6}$ with a linear warmup on the first 30\% of the training steps from $0$ to $\eta$.
Every 40 updates of the gradients, the models were evaluated on the early stopping data by calculating the macro-F1 score.
If the score did not increase for two consecutive evaluations the training was interrupted and the model achieving the largest F1 score on the early stopping set was used for evaluation on the validation fold or test data.

\emph{Loss functions.} When training single-label models, we used a negative log-likelihood loss function.
Multi-label models were trained by minimizing the binary cross entropy loss function averaged over the three classes for every sample in a mini-batch.

\emph{Threshold selection.} In multi-label models, a sample (comment) from the dataset was predicted to belong to those classes for which the respective class probabilities of the model exceeded a certain threshold.
Since multi-label models can attribute a sample to three classes, three different thresholds needed to be determined.
We chose these thresholds by evaluating the model for threshold values between 0 and 1 (exploring this range with a step size of 0.025) on the data reserved for early stopping and accepting the values achieving the highest F1 scores for each class separately.
In single-label models, we did not need to chose any thresholds since the class membership of a sample was predicted by identifying the largest output probability of the two output neurons.

\subsection{Ensembling}
\label{ssec:ensembling}

Training complex models such as GBERT or GELECTRA on small datasets can lead to overfitting.
Following the work by \citet{Risch2020}, we counteracted this phenomenon by creating ensembles of models using bootstrap aggregation.
Ensemble members differed in the initial weights of the classification layers and the data samples randomly selected for early stopping.
The predictions of an ensemble were determined by averaging the predicted probabilities of the ensemble members (\emph{soft majority voting}).
In single-label models, a model's prediction was then determined by identifying the output neuron associated with the largest ensemble-averaged output probability.
In multi-label models, a model predicted a sample to belong to certain classes if ensemble-averaged class probabilities exceeded optimal thresholds.
The optimal thresholds were determined by evaluating each ensemble member for all thresholds on the early stopping data (see section~\ref{sec:training}) and accepting the thresholds with the highest macro-F1 score as the optimal values.

\section{Results}

\begin{table*}
    \begin{center}
      \begin{tabular}{c|c|c|c|c}
          & $\mathbf{\overline{F1}}$ & $\mathbf{F1_{toxic}}$ & $\mathbf{F1_{engaging}}$ & $\mathbf{F1_{fact}}$\\
          \hline
          \multicolumn{5}{c}{\textit{model exploration}}\\
          \hline

          50 GELECTRA & & & & \\
          multi-label & 0.765 (0.008) & 0.730 (0.018) & 0.782 (0.018) & 0.784 (0.019) \\
          \hline

          50 GBERT & & & & \\
          multi-label & 0.760 (0.002) & 0.720 (0.006) & 0.777 (0.015) & 0.782 (0.013) \\
          \hline

          25+25 GELECTRA/GBERT & & & & \\
          multi-label & 0.763 (0.007) & 0.726 (0.010) & 0.780 (0.015) & 0.784 (0.015) \\
          \hline

          25+25 GELECTRA/GBERT & & & & \\
          single-label & 0.768 (0.006) & 0.736 (0.011) & 0.782 (0.014) & 0.787 (0.013) \\
          \hline
          \hline
          \multicolumn{5}{c}{\textit{final submissions}}\\
          \hline

          200 GELECTRA & & & & \\
          multi-label & 0.717 & 0.713 & 0.690 & 0.748 \\
          \hline

          200+200 GELECTRA/GBERT & & & & \\
          multi-label & \textbf{0.726} & 0.716 & \textbf{0.699} & \textbf{0.763} \\
          \hline

          30+30 GELECTRA/GBERT & & & & \\
          single-label & 0.699 & \textbf{0.718} & 0.658 & 0.723 \\
          corrected scores & \textbf{0.727} & \textbf{0.717} & 0.697 & \textbf{0.768} \\
      \end{tabular}
    \end{center}
    \caption{
        F1 scores achieved by different ensembles during model exploration on the validation folds (rows 1--4; mean and standard deviation over the folds) and F1 scores achieved by the submitted models on the test data as reported by the GermEval 2021 organizers (rows 5--7; best scores are shown in bold).
        The \emph{corrected scores} shown in the last row were calculated after correcting an error identified after submission.
      }
      \label{tab:f1scores}
\end{table*}

\begin{figure}
    \includegraphics[width=\linewidth]{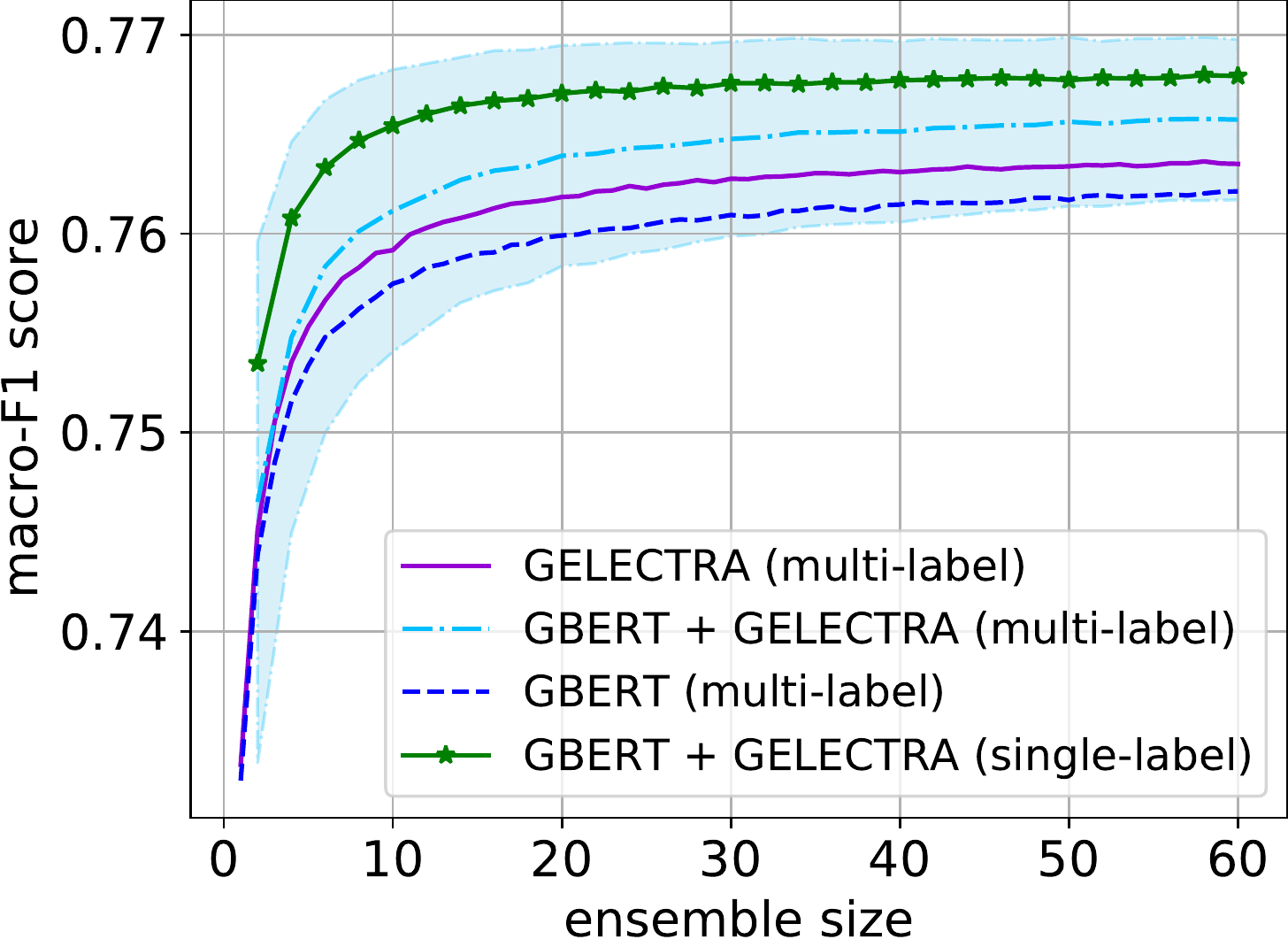}
    \caption{Dependence of the average macro-F1 score (lines) on ensemble size for different ensemble compositions. Standard deviations are shown as blue shaded area for the multi-label ensemble GBERT/GELECTRA. Note that for ensemble sizes larger than 30, average macro-F1 scores differed between ensembles only in their third decimal place, a variation that we considered insignificant.}
    \label{fig:ensemble}
\end{figure}

\emph{Model exploration.}
We investigated whether and how classification performance (quantified by macro-F1 score) depended on (i) ensemble size, (ii) ensemble composition, and (iii) on whether ensemble members can exploit label correlations (\mbox{multi-label} models) or not (single-label models). To study the effect of (ii), we compared the classification performance of different ensemble compositions. The first ensemble consisted of fine-tuned multi-label GELECTRA models, while the second ensemble consisted of fine-tuned multi-label GBERT models. In a third ensemble we used equal parts of fine-tuned multi-label GELECTRA and GBERT models. To study the effect of (iii), we compared the third ensemble with a fourth ensemble which was composed of equal parts of \mbox{fine-tuned} single-label GELECTRA and GBERT models. Finally, we investigated the effect of (i) via a bootstrap experiment following \citet{Risch2020}.

The bootstrap experiment was carried out using a 5-fold cross validation scheme. We trained 100 models each of multi-label GBERT and multi-label GELECTRA, and 50 models each of single-label GBERT and single-label GELECTRA on each cross-validation split. For a given ensemble size, we created 1000 ensembles by randomly sampling with replacement from the set of trained models. Each ensemble made predictions on a validation fold by \emph{soft majority voting}. The average macro-F1 score of an ensemble was determined by averaging the macro-F1 scores obtained on each of the 5 validation folds. Thus, for a given ensemble size, we obtained 1000 average macro-F1 scores.

Figure~\ref{fig:ensemble} shows the mean of the average macro-F1 scores obtained for different ensemble sizes and ensemble compositions. We observed classification performance to increase with ensemble size, irrespective of model composition and of whether models could or could not exploit label correlations. Largest increases were found for ensemble sizes up to 15 ensemble members, which is consistent with a previous study on a different classification task \citep{Risch2020}. Moreover, macro-F1 scores continued to increase beyond the ensemble size of 15.

For a given ensemble size larger than 30, classification performance between ensembles of the different compositions varied only in the third decimal of their macro-F1 score, a variation that we did not consider significant. Ensembles consisting of 100\% GELECTRA models, 100\% GBERT models, or 50\% GELECTRA and 50\% GBERT models yielded comparable macro-F1 scores. Likewise, ensembles consisting of either multi-label or \mbox{single-label} models showed comparable macro-F1 scores for a fixed ensemble size. These observations were confirmed by F1 scores obtained for ensembles of size 50, reported in table~\ref{tab:f1scores} (rows 1--4).

\emph{Submitted models.} Three ensembles were submitted and evaluated on the test data of the shared tasks reflecting the lines of investigation laid out before. The evaluated ensembles were (1) an ensemble of 200 multi-label GELECTRA models, (2) an ensemble of 200 multi-label GELECTRA and 200 multi-label GBERT models, and (3) an ensemble of 30 single-label GELECTRA and 30 single-label GBERT models which were trained on all the training data (see section~\ref{sec:training}). We note that time and computational constraints limited ensemble sizes.

On the test data of the shared task, ensemble (2) achieved the largest macro-F1 score of $0.73$, followed by ensemble (1) with $0.72$ and (3) with $0.70$ (see table~\ref{tab:f1scores}, rows 5--7). We identified a software bug after submission deadline that affected the scores calculated for ensemble (3) which achieved a corrected macro-F1 score of $0.73$. These results supported observations made during model exploration that ensemble composition and classification type did not significantly affect classification performance for ensemble sizes larger than 30.

\section{Conclusion}
We trained ensembles of fine-tuned German language models, namely GELECTRA and GBERT, to classify German
toxic, engaging, and \mbox{fact-claiming} comments in the GermEval 2021 shared task. We investigated whether classification performance (quantified by macro-F1 scores) depended on (i) ensemble size, (ii) ensemble composition, or (iii) whether models were trained as multi-label classifiers (and thus potentially exploiting label correlations) or as single-label classifiers. We observed that ensemble size had a significant effect on classification performance, with more ensemble members leading to better macro-F1 scores, consistent with previous observations by \citet{Risch2020} on a different dataset. Neither ensemble composition nor model classification type (multi- or single-label) showed significant different classification performance for the studied parameters when the ensemble size was larger than 30. Two ensembles achieved the largest macro-F1 score ($0.73$) on the test data, namely the multi-label and single-label ensembles consisting of GELECTRA and GBERT models.

\appendix

\end{document}